\documentclass[journal,11pt,draftclsnofoot,onecolumn]{IEEEtran}
\usepackage{amsmath,amsfonts}
\usepackage{algorithmic}
\usepackage{array}
\usepackage[caption=false,font=normalsize,labelfont=sf,textfont=sf]{subfig}
\usepackage{textcomp}
\usepackage{stfloats}
\usepackage{url}
\usepackage{verbatim}
\usepackage{graphicx}

\usepackage{threeparttable}
\usepackage{multirow}
\usepackage{multicol}
\usepackage{makecell}
\usepackage{booktabs} 
\usepackage{amsmath}
\usepackage{amssymb}
\usepackage{amsthm}
\usepackage{hyperref} 

\usepackage[T1]{fontenc}
\usepackage[utf8]{inputenc}
\usepackage{authblk}

\usepackage{graphicx}
\usepackage{caption}
\usepackage{lipsum}
\usepackage{xcolor}
\usepackage{color, colortbl}
\definecolor{cgreen}{RGB}{11,135,28}

\usepackage{amsmath}
\usepackage{amsfonts}

\newcommand{\Jiuniu}[1]{\textcolor{red}{[Jiuniu: #1]}}

\def\BibTeX{{\rm B\kern-.05em{\sc i\kern-.025em b}\kern-.08em
    T\kern-.1667em\lower.7ex\hbox{E}\kern-.125emX}}
\usepackage{balance}

\title{UAV-VisLoc: A Large-scale Dataset for UAV Visual Localization}
\author[1 *]{Wenjia Xu\thanks{* Equal contribution}\thanks{Corresponding author: Wenjia Xu (xuwenjia@bupt.edu.cn)}}
\author[1 *]{Yaxuan Yao}
\author[1 *]{Jiaqi Cao}
\author[2]{Zhiwei Wei}
\author[2]{Chunbo Liu}
\author[3]{\\Jiuniu Wang}
\author[1]{Mugen Peng}

\affil[1]{\centering {State Key Laboratory of Networking and Switching Technology,} \newline \centering {Beijing University of Posts and Telecommunications}}
\affil[2]{Aerospace Information Research Institute, Chinese Academy of Sciences}
\affil[3]{City University of Hong Kong}

\begin{document}
\maketitle

\begin{abstract}
The application of unmanned aerial vehicles~(UAV) has been widely extended recently. It is crucial to ensure accurate latitude and longitude coordinates for UAVs, especially when the global navigation satellite systems (GNSS) are disrupted and unreliable. Existing visual localization methods achieve autonomous visual localization without error accumulation by matching the ground-down view image of UAV with the ortho satellite maps. However, collecting UAV ground-down view images across diverse locations is costly, leading to a scarcity of large-scale datasets for real-world scenarios. Existing datasets for UAV visual localization are often limited to small geographic areas or are focused only on urban regions with distinct textures. 
To address this, we define the UAV visual localization task by determining the UAV's real position coordinates on a large-scale satellite map based on the captured ground-down view. 
In this paper, we present a large-scale dataset, UAV-VisLoc, to facilitate the UAV visual localization task. This dataset comprises images from diverse drones across 11 locations in China, capturing a range of topographical features.
The dataset features images from fixed-wing drones and multi-terrain drones, captured at different altitudes and orientations. 
Our dataset includes 6,742 drone images and 11 satellite maps, with metadata such as latitude, longitude, altitude, and capture date. Our dataset is tailored to support both the training and testing of models by providing a diverse and extensive data. Our dataset is released at \textit{\href{https://github.com/IntelliSensing/UAV-VisLoc}{https://github.com/IntelliSensing/UAV-VisLoc}}

\end{abstract}

\begin{IEEEkeywords}
Unmanned Aerial Vehicle, Visual Localization, UAV-VisLoc Dataset
\end{IEEEkeywords}

\section{Introduction}
The utilization of UAVs has expanded significantly across various fields recently, such as agriculture development, environment monitoring, and ground surveillance. Acquiring the UAV's precise location (i.e., accurate latitude and longitude coordinates) is crucial for the downstream tasks, especially under conditions where Global Navigation Satellite Systems (GNSS) are interrupted or unavailable. Meanwhile, with the development of remote sensing technology, we can obtain high-resolution satellite maps covering almost everywhere on Earth, with each pixel annotated with precise coordinates. 
Therefore, matching the ground-down view captured by drone images with the corresponding ortho-satellite imagery allows accurate and quick UAV localization with no error accumulation~\cite{dissanayake2001solution,lu2018survey,li2023jointly}.

The UAV visual localization task, which locates the UAV based on the captured ground-down view, has gained much attention in recent years.  ASIFT~\cite{wang2018unmanned} obtains UAV coordinates by matching oblique UAV down-views with remote sensing images. In addition, LCM~\cite{ding2020practical} and GLVL~\cite{li2023jointly} obtain UAV coordinates by matching orthorectified UAV down-views with remote sensing images.
Wang et al.~\cite{wang2018unmanned} proposed ASIFT as an affine scale-invariant feature transform-based method that can register UAV oblique images at a subpixel level.
Ding et al.~\cite{ding2020practical} proposed a cross-view matching method LCM based on location classification, which improved accuracy by simplifying the retrieval problem into a classification problem and considering the effect of feature vector size.
Li et al.~\cite{li2023jointly} proposed a two-stage visual localization method GLVL, which combines a large-scale retrieval module for finding similar regions and a fine-grained matching module to locate UAV coordinates precisely.

The dataset benchmarks are necessary when evaluating previous methods with a uniform standard, which is shown in TABLE~\ref{tab:UAV_datasets}. 
The CVUSA~\cite{workman2015wide} defined a ground-satellite matching task, which provides a dataset involving street-view images and satellite-view images captured from distinct regions across the United States. 
VIGOR~\cite{zhu2021vigor} redefined the problem by breaking the one-to-one retrieval setup of the dataset, so that there are often multiple reference satellite map blocks covering a single query location. 
The University-1652~\cite{zheng2020university} introduced drone-view to cross-view geo-localization, extended the visual localization task as ground-drone-satellite cross-view matching, and provided a three-view dataset from 1,652 architectures of 72 universities.
Furthermore, the SUE-200~\cite{zhu2023sues} provided drone images from multi-scene at multi-height for each satellite map. 
Recently, the DenseUAV~\cite{dai2022vision} focused on the problem of UAV self-localization, considering both prominent features and the spatial distribution of objects in images.

\begin{table*}[t]
	\centering
	\renewcommand{\arraystretch}{1}
	\caption{A SUMMARY OF RELATED UAV VISUAL LOCALIZATION DATASETS.}
	\begin{threeparttable}
 \resizebox{\linewidth}{!}{
	\begin{tabular}{cccc}
		\cline{1-4}
        \Xhline{1.2pt}
		\textbf {Dataset} & \textbf {Image Size}& \textbf {Region} & \textbf {Platform}  \\
        \cline{1-4}
		\textbf {CVUSA~\cite{workman2015wide}} &  {71 k}& {United States} & {Ground-Satellite}     \\
		\textbf {VIGOR~\cite{zhu2021vigor}} &  {144 k}& {Manhattan, San Francisco, Chicago, and Seattle} & {Ground-Satellite}   \\
		\textbf {University-1652~\cite{zheng2020university}} &  {50.2 k} & {1,652 architectures of 72 universities} & {Ground-Drone-Satellite}   \\
		\textbf {SUE-200~\cite{zhu2023sues}}  &{6.1 k}& {a university in Shanghai, China} & {Drone-Satellite}  \\
		\textbf {DenseUAV~\cite{dai2022vision}}  &{20.3 k} & {14 universities in Zhejiang, China} & {Drone-Satellite}  \\
		\Xhline{1.2pt}
	\end{tabular}}
        \end{threeparttable}
	\label{tab:UAV_datasets}
\end{table*}

However, the existing Cross-View Image Geo-localization datasets can not meet the practical requirements and suffer from the following shortcomings.
\textbf{(i)} Simple Task. CVUSA~\cite{workman2015wide} and VIGOR~\cite{zhu2021vigor} was initially proposed to solve the ground-to-aerial matching problem.
The DenseUAV~\cite{dai2022vision} simply considers the visual localization task as a retrieval task. This dataset uses blocks of satellite maps that are based solely on the coordinates of the query images. It is different from the real scenario, which locates the latitude and longitude of a UAV on a satellite map.
\textbf{(ii)} Low Altitude. Most existing drone images of realistic scenes are of urban areas captured by multi-rotor drones, which often fly at a low altitude, e.g., University-1652~\cite{zheng2020university} and SUE-200~\cite{zhu2023sues} use sampling altitudes of no more than 300 meters. We can not obtain a general visual localization model only with the data collected from the low altitude. 

\textbf{(iii)} Single Scene. University-1652~\cite{zheng2020university}, SUE-200~\cite{zhu2023sues} and DenseUAV~\cite{dai2022vision} drone images are only of texture-rich campuses or urban areas with a single location and a single geographic feature. The application scenarios are limited for the models trained with the single scene dataset.

\captionsetup[figure]{name={Fig.},labelsep=period}
\begin{figure}[!t]
\centering
\captionsetup{font={footnotesize,stretch=1.25}}
\includegraphics[width=1\textwidth]{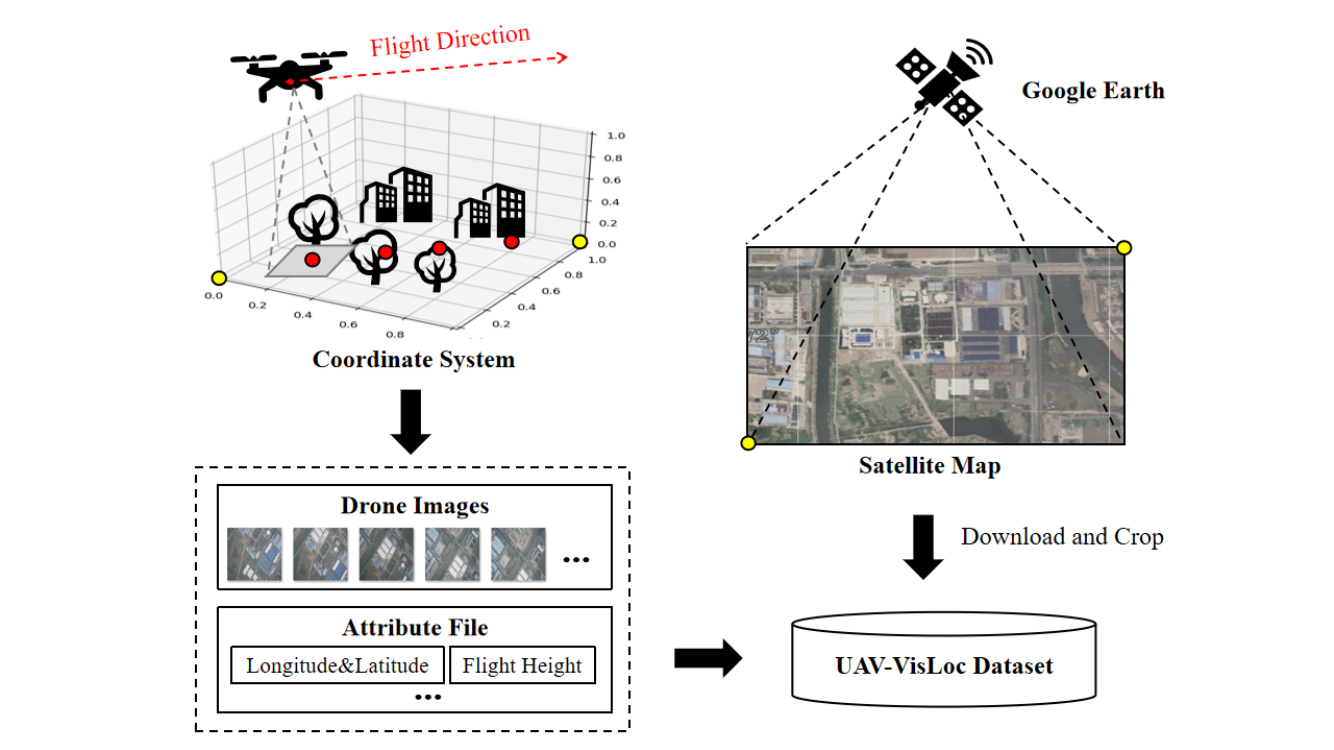}
\caption{The dataset collecting process. The red point in the coordinate system represents the projection of the drone's current location on the ground, i.e., the center point of the image taken by the drone. The yellow points represent the satellite map boundaries of the entire flight range.  }
\label{fig:fig1}
\end{figure}

In summary, we unify the definition of visual localization methods for UAVs in real scenarios. We propose a dataset as a benchmark for the visual localization task. The dataset collecting process is shown in Fig.~\ref{fig:fig1}. The task is defined as a UAV obtaining the coordinates of its current location by matching the down image with the satellite map when losing its GNSS coordinates. Thus, we collect a sequence of images on the UAV flight track in our dataset. And we further collect an orthorectified remote sensing map covering a large geographic area where every pixel is labelled with coordinates. 
The contributions of this dataset are as follows:

\begin{itemize}

\item{\textbf{Fixed-wing drone images.} Fixed-wing UAVs have endurance and high flight altitude and are now widely used in surveying and mapping, agriculture, and reconnaissance industries. Therefore, in addition to multi-rotor UAVs, our dataset is supplemented with drone images captured at higher altitudes by fixed-wing UAVs to accommodate multi-domain model training and testing.}

\item{\textbf{Multi-terrain drone images.} Our dataset contains drone images of various geomorphological texture features, such as villages, towns, farms, cities, rivers, hills, etc., covering most of China.}

\item{\textbf{Multi-height and multi-heading-angle drone images.} Our dataset contains drone images at different altitudes, with both low-altitude urban scenes and high-altitude field scenes. Our dataset also contains flight heading angle information, which benefits subsequent related studies.}

\item{\textbf{Large-scale visual localization dataset.} This dataset provides a sufficient amount of data to support the training and testing of models. In detail, it has 6,742 drone images and 11 satellite maps.}
\end{itemize}

\section{Data collection}
\subsection{Drone Images}
For the UAV visual localization task in real scenario, we collected drone images from different locations within China, such as Jiangsu Province, Taihu Lake Coast, Yangtze River Basin, Donghuayuan Town, Yunnan Province, and Gansu Province. The raw drone image data are characterized as follows. 
{\bf (i)} The drone images cover a wide range of complex topographical and geomorphological features, such as villages, towns, farms, cities, rivers, hills, and forests.
{\bf (ii)} We capture drone images at different heights, from 400 meters to 2,000 meters, with different heading angles. We use both multi-rotor and fixed-wing drones.
{\bf (iii)} The drone images we captured contain two seasons, i.e., summer and autumn.

\subsection{Satellite Maps}
\textbf{Satellite Map Sources.}
Satellite maps are all taken from Google Earth, with a spatial resolution of 0.3 meters.

\textbf{Download and Crop.}
A satellite map covering all ranges captured by the lower view of the UAV is downloaded separately for each UAV image of each flight. The data can be further preprocessed according to different training or testing tasks.

\begin{table*}[t]
	\centering
	\renewcommand{\arraystretch}{1} 
	\caption{IMAGE ATTRIBUTE COMPOSITION FOR THE UAV-VisLoc DATASET.}
	\begin{threeparttable}
 \resizebox{\linewidth}{!}{
	\begin{tabular}{c|cccc|ccc}
		\cline{1-8}
        \Xhline{1.2pt}
		\multirow{2}{*}{\textbf{Site}} &\multicolumn{4}{c|}{\rule{0pt}{10pt}\textbf{Drone Image}} & \multicolumn{3}{c}{\textbf{Satellite Map}}\\
		\cline{2-8}
		\multicolumn{1}{c|}{}  & \textbf {Pixel} & \textbf {Date}& \textbf {Phi} \tnote{*}  & \textbf {Height}  & \textbf {Pixel} & \textbf {Date}& \textbf {Coordinates Range} \tnote{**}   \\
\cline{1-8}
        \multirow{2}{*}{Changjiang-20} & \multirow{2}{*}{3976×2652} & \multirow{2}{*}{2018-09}&\multirow{2}{*}{165°}& \multirow{2}{*}{405 m} &\multirow{2}{*}{9774×26762}& \multirow{2}{*}{2023-11} &{29.774065°N, 115.970635°E}\\
        \multicolumn{1}{c|}{} & {} & {} & {} & {} & {} &  {} & {29.702283°N, 115.996851°E}\\
        \cline{1-8}
		\multirow{2}{*}{Changjiang-23} & \multirow{2}{*}{3976×2652} & \multirow{2}{*}{2018-09}& \multirow{2}{*}{5°}& \multirow{2}{*}{405 m} &\multirow{2}{*}{11482×34291}& \multirow{2}{*}{2022-09} &{29.817376°N, 116.033769°E}\\
        \multicolumn{1}{c|}{} & {} & {} & {} & {} & {} & {}& {29.725402°N, 116.064566°E}\\
		\cline{1-8}
		\multirow{2}{*}{Taizhou-1} &\multirow{2}{*}{3976×2652} 	& 	\multirow{2}{*}{2018-10}&\multirow{2}{*}{-40°}&\multirow{2}{*}{466 m} 	&\multirow{2}{*}{35092×24308}&\multirow{2}{*} {2021-04} & {32.355491°N, 119.805926°E}\\
        \multicolumn{1}{c|}{} & {} & {} & {} & {} & {} &  {}& {32.290290°N, 119.900052°E}	\\
		\cline{1-8}
		\multirow{2}{*}{Taizhou-6} &\multirow{2}{*}{3976×2652} & 	\multirow{2}{*}{2018-10}& \multirow{2}{*}{170°}&\multirow{2}{*}{542 m}	&\multirow{2}{*}{18093×38408}& \multirow{2}{*}{2023-03}&{32.254036°N, 119.90598°E}\\
        \multicolumn{1}{c|}{} & {} & {} & {} & {} & {} &  {}& {32.151018°N, 119.954509°E}	\\
		\cline{1-8}
		\multirow{2}{*}{Yunnan} &\multirow{2}{*}{3000×2000} & 	\multirow{2}{*}{2016-06}&\multirow{2}{*}{100°}&\multirow{2}{*}{/}	&\multirow{2}{*}{9394×6144}&\multirow{2}{*} {2022-03} &{24.666899°N, 102.340055°E}\\
        \multicolumn{1}{c|}{} & {} & {} & {} & {} & {} &  {}& {24.650422°N, 102.365252°E} \\
		\cline{1-8}
		\multirow{2}{*}{Zhuxi} &\multirow{2}{*}{3976×2652} & \multirow{2}{*}{2022-10}&\multirow{2}{*}{-15°}& \multirow{2}{*}{840 m} &\multirow{2}{*}{4119×8592}& \multirow{2}{*}{2022-09} &{32.373177°N, 109.635160°E}\\
        \multicolumn{1}{c|}{} & {} & {} & {} & {} & {} &  {}& {32.346944°N, 109.656837°E} \\
		\cline{1-8}
		\multirow{2}{*}{Donghuayuan} &\multirow{2}{*}{3000×2000} 	& 	\multirow{2}{*}{2018-07}&\multirow{2}{*}{-1.5°}& \multirow{2}{*}{688 m}	&\multirow{2}{*}{3000×170}& \multirow{2}{*}{2023-06}	&{40.340058°N, 115.791182°E}\\
		\multicolumn{1}{c|}{} & {} & {} & {} & {} & {} &  {}& {40.339604°N, 115.799230°E}	\\
		\cline{1-8}
        \multirow{2}{*}{Huzhou-3} &\multirow{2}{*}{3976×2652} & 	\multirow{2}{*}{2019-06}&\multirow{2}{*}{100°}&\multirow{2}{*}{551 m}  &\multirow{2}{*}{43421×16294}& \multirow{2}{*}{2023-07} &{30.947227°N, 120.136489°E}	\\
        \multicolumn{1}{c|}{} & {} & {} & {} & {} & {} & {}& {30.903521°N, 120.252951°E}	\\
		\cline{1-8}
        \multirow{2}{*}{Huzhou-6} &\multirow{2}{*}{3976×2652} 	& 	\multirow{2}{*}{2019-06}&\multirow{2}{*}{-50°}&\multirow{2}{*}{546 m} &\multirow{2}{*}{44800×33280}& \multirow{2}{*}{2024-01} &{30.999512°N, 120.030076°E}	\\
        \multicolumn{1}{c|}{} & {} & {} & {} & {} & {} &  {}& {30.910733°N, 120.149648°E}	\\
		\cline{1-8}
		\multirow{2}{*}{Huailai} &\multirow{2}{*}{3000×2000} 	& 	\multirow{2}{*}{2018-09}&\multirow{2}{*}{170°}& \multirow{2}{*}{772 m} &\multirow{2}{*}{6593×5077}& \multirow{2}{*}{2023-06}	&{40.355093°N, 115.776356°E}\\
        \multicolumn{1}{c|}{} & {} & {} & {} & {} & {} &  {}& {40.341475°N, 115.794041°E}\\
		\cline{1-8}
		\multirow{2}{*}{Shandan}	&\multirow{2}{*}{3976×2652} 	& 	\multirow{2}{*}{2023-10}&\multirow{2}{*}{90°}&\multirow{2}{*}{/} 	&\multirow{2}{*}{29592×16582}& \multirow{2}{*}{2021-03}	&{38.852301°N, 101.013109°E}\\
		\multicolumn{1}{c|}{} & {} & {} & {} & {} & {} &  {}& {38.807825°N, 101.092483°E}	\\
		\cline{1-8}
		\Xhline{1.2pt}
	\end{tabular}}
        \begin{tablenotes}
            \footnotesize 
            \item[*] Phi denotes the drone heading angle. The drone heading angle is the angle between the UAV longitudinal axis and the Earth's North Pole and takes values in the range [-180°, 180°]. If the flight direction is due east, the Phi is 0°. If the flight direction is north, Phi is positive, and if the flight direction is south, Phi is negative. 
            \item[**] Coordinates Range denotes the latitude and longitude range covered by satellite maps.
            
            \end{tablenotes} 
        \end{threeparttable}
	\label{tab:image_attribute}

\end{table*}

\captionsetup[figure]{name={Fig.},labelsep=period}
\begin{figure}[!t]
\centering
\captionsetup{font={footnotesize,stretch=1.25}}
\includegraphics[width=1\textwidth]{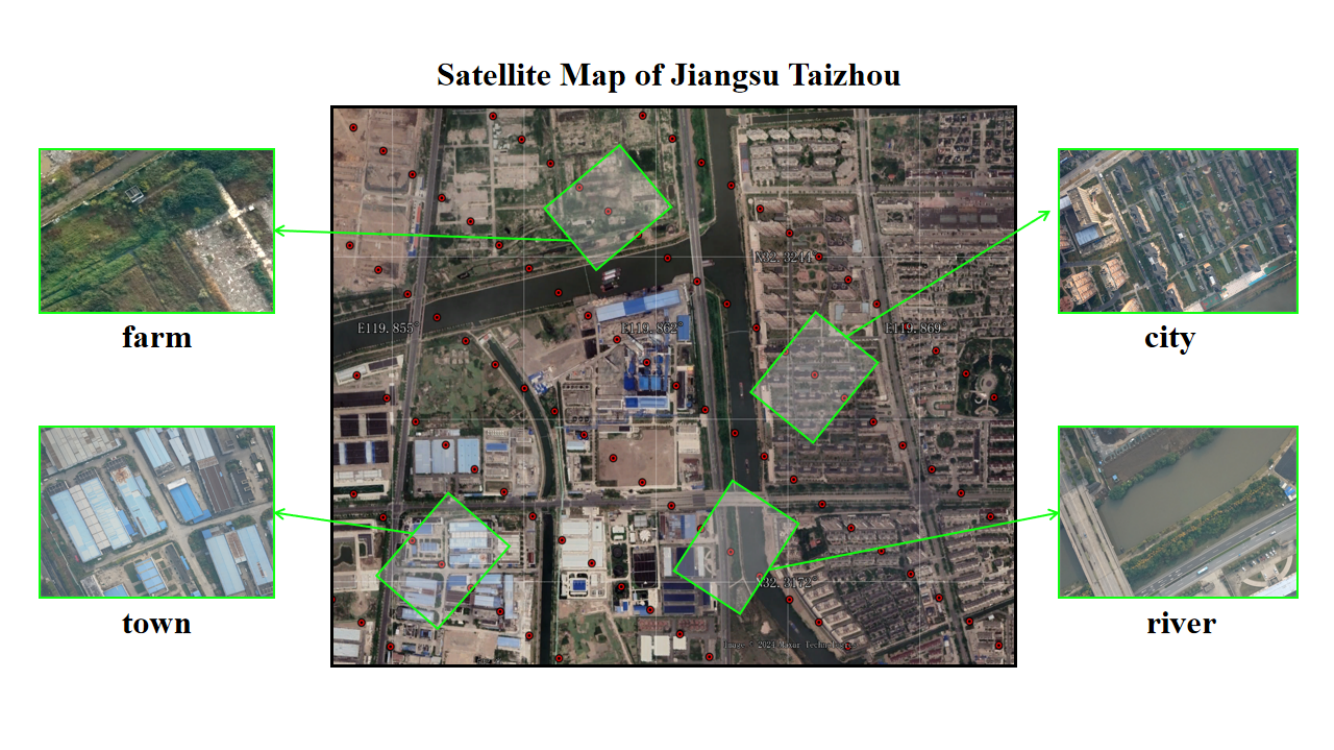} 
\caption{An example of drone images and satellite map. The red dots in satellite map represent the center points of drone images. The satellite map encompasses various terrains such as cities, towns, farms, and rivers. We also show the drone images of these terrains.} 
\label{fig:fig2}
\end{figure}

\section{Dataset Construction}
Our dataset contains 6,742 drone images and 11 satellite maps. As presented in TABLE~\ref{tab:image_attribute}, we provide the attributes of drone images, such as latitude and longitude of the center point, shooting height, shooting date, and heading angle (Phi). We also provide GPS latitude and longitude ranges for 11 satellite maps.
As shown in Fig.~\ref{fig:fig2}, each satellite map downloaded from Google Maps contains all UAV images captured by a single drone flight, which may encompass various terrains such as cities, towns, farms, and rivers. 

\section{Future Perspectives}

Our UAV-VisLoc dataset provides training and testing samples for UAV visual localization. The visual localization task aims to find the most similar satellite map to localize the drone's latitude and longitude coordinates within the satellite map. Additionally, the drone intends to find the most relevant place (drone images) by where it has passed. According to its flight history, the drone could be navigated back to the target place.

\newpage
\bibliographystyle{IEEEtran}
\bibliography{ref}

\end{document}